\documentclass[10pt,twocolumn,letterpaper]{article}

\usepackage[pagenumbers]{wacv} 

\definecolor{wacvblue}{rgb}{0.21,0.49,0.74}
\usepackage[pagebackref,breaklinks,colorlinks,allcolors=wacvblue]{hyperref}

\def\wacvPaperID{1557} 
\def\confName{WACV}
\def\confYear{2026}

\title{SAFER-AiD: Saccade-Assisted Foveal-peripheral vision Enhanced Reconstruction for Adversarial Defense}

\author{Jiayang Liu${^1}$, Daniel Ts'o${^2}$, Yiming Bu${^1}$, Qinru Qiu${^1}$\\${^1}$Department of Electrical Engineering and Computer Science, Syracuse University\\${^2}$ SUNY Upstate Medical University\\ \{jliu206,\ ybu104,\ qiqiu\}@syr.edu,\           \{tsod\}@upstate.edu
}

\begin{document}
\maketitle
\begin{abstract}
Adversarial attacks significantly challenge the safe deployment of deep learning models, particularly in real-world applications. Traditional defenses often rely on computationally intensive optimization (e.g., adversarial training or data augmentation) to improve robustness, whereas the human visual system achieves inherent robustness to adversarial perturbations through evolved biological mechanisms. We hypothesize that attention guided non-homogeneous sparse sampling and predictive coding plays a key role in this robustness. To test this hypothesis, we propose a novel defense framework incorporating three key biological mechanisms: foveal-peripheral processing, saccadic eye movements, and cortical filling-in. Our approach employs reinforcement learning-guided saccades to selectively capture multiple foveal-peripheral glimpses, which are integrated into a reconstructed image before classification. This biologically inspired preprocessing effectively mitigates adversarial noise, preserves semantic integrity, and notably requires no retraining or fine-tuning of downstream classifiers, enabling seamless integration with existing systems. Experiments on the ImageNet dataset demonstrate that our method improves system robustness across diverse classifiers and attack types, while significantly reducing training overhead compared to both biologically and non-biologically inspired defense techniques.
\end{abstract}
    
\section{Introduction}
\label{sec:intro}
The transition from laboratory-developed AI models to real-world deployment, especially safety-critical domains (e.g., medical diagnostics, autonomous driving), faces a critical hurdle: the vulnerability of models to subtle distribution shifts and malicious inputs. Among these threats, adversarial attacks pose a particularly insidious risk. Even imperceptible perturbations (e.g., $\epsilon \leq 8/255$ in pixel value) can catastrophically degrade model performance, undermining trust in AI systems.

Adversarial training (AT) \cite{AT2} has become the gold standard for improving model robustness to white-box attacks by optimizing against worst-case perturbations during training. However, this approach is known for its high computational cost\cite{AT1}. The training complexity becomes even more prohibitive if comprehensive robustness against diverse attack strategies is pursued, as adversarial training must be performed for each potential attack model. Compared to white-box attacks where attackers are assumed to have access to the target model's parameters or architecture, transferable attacks, where adversarial examples are crafted using a surrogate model, represent a more practical threat \citep{APGD,Deepfool,MIFGSM,TGR}. The surrogate model can be any commonly used network architecture, such as VGGs\citep{VGG}, ViTs\citep{ViT} and ResNets\citep{resnet}. In the absence of knowledge about the surrogate model, adversarial training often falls short against transferable attacks. To address this limitation, Ensemble Adversarial Training \citep{EnsembleAT} has been proposed, where multiple models with diverse architectures are trained against a range of white-box attacks. However, this technology further increases the computational burden, highlighting a  trade-off between robustness and efficiency.

Biological vision systems exhibit inherent robustness to adversarial perturbations, inspiring a line of research into biologically inspired input transformations \citep{rblur,rwrap,FTT,FoveaT,findingrobustfeat}. These approaches aim to enhance model robustness while reducing the dependence on computationally expensive adversarial training. However, a notable performance gap still remains, particularly under strong attacks. Moreover, these methods require training a new classifier to process the transformed inputs, which demands substantial labeled data. Due to information loss introduced by the transformation, they also tend to degrade accuracy on clean (non-adversarial) images.  Notably, most existing work focuses primarily on the foveation mechanism, overlooking other key components of biological vision, such as attention-guided saccades and cortical filling-in, that also contribute to human visual robustness. 

In this work, we aim to bridge this gap by designing a defense framework rooted in the hypothesis that attention guided non-homogeneous sparse sampling and predictive reconstruction are key contributors to the robustness of biological perception. The proposed Saccade-Assisted Foveal-peripheral vision Enhanced Reconstruction for Adversarial Defense(SAFER-AiD) framework first applies foveal-peripheral sampling to suppress adversarial perturbations in less informative regions while preserving semantic coherent in salient areas. A reinforcement learning-guided saccadic mechanism then sequentially selects informative regions of the image to attend to, mimicking how the human eye actively explores a scene. Finally, a reconstruction model, guided by the stream of foveal-peripheral inputs, fills in missing information to synthesize a complete scene representation. The  reconstruction removes adversarial perturbations while maintaining high fidelity to the original clean image. 

SAFER-AiD is classifier-agnostic and functions as a plug-and-play preprocessing module that can be inserted before any classifier, offering a countermeasure against a wide range of adversarial attacks. By sparsely sampling the peripheral regions, it effectively removes adversarial perturbed pixels introduced by attackers. The reconstruction process then fills in missing pixels and replaces corrupted ones using prior knowledge learned from clean images. To compensate for information loss due to sparse sampling, the saccade mechanism dynamically positions the foveal center to capture the most informative regions at full resolution. Unlike traditional adversarial training, which must be tailored to each specific attack model, SAFER-AiD employs a fundamentally different defense strategy and requires only a single, unsupervised training phase. As a result, it significantly reduces overall training complexity.     

To emphasize the practical utility of our approach, we focus on black-box attack scenarios, including transferable and gradient-free attacks. We evaluate the SAFER-AiD with a diverse set of classifiers, including eight Vision Transformer (ViT) variants and 3 convolutional neural network (CNN) architectures on the ImageNet dataset. Our method achieves at least 13.3\% increase in top-1 accuracy under Token Gradient Regularization (TGR) attacks and improvements ranging from 9.6\% to 12.9\% under Momentum Iterative Fast Gradient Sign Method (MI-FGSM) attacks. Against gradient-free attacks, SAFER-AiD yields a 15.2\% increase in top-1 accuracy under attack budge $\epsilon = 16/255$ across different classifiers. Contributions of this work can be summarized as follows:

\begin{itemize}
    \setlength\itemsep{2pt}
    \setlength\topsep{2pt}
    \item \textbf{Biologically inspired design:} We present the first general purpose adversarial defense framework, SAFER-AiD, that systematically integrates the core biological mechanisms underlying human visual robustness—\textit{foveal-peripheral sampling}, \textit{saccadic eye movements}, and \textit{cortical filling-in}. This design is rooted in neuroscience and validated through extensive empirical evaluation.
    
    \item \textbf{Superior robustness performance:} SAFER-AiD outperforms state-of-the-art biologically inspired defenses, achieving a 11.9\% improvement under $\epsilon = 2/255$, and retains robustness even under stronger attacks where existing baselines fail completely .
    
    \item \textbf{Significantly lower training cost:} In comparison to fast adversarial training, our method is substantially more training-efficient. It delivers at least 5.8\% higher top-1 accuracy while requiring only a single NVIDIA RTX 2080 GPU, whereas adversarial training baselines demand 4$\times$P100 GPUs and significantly longer runtimes. Additionally, adversarial training must be repeated from scratch whenever the classifier architecture changes, while SAFER-AiD is agnostic to the architecture of downstream classifier.
    
    \item \textbf{Model-agnostic robustness:} The SAFER-AiD framework demonstrates consistent improvements in robustness across diverse model architectures, including both CNNs and Vision Transformers, while preserving high accuracy on clean images.
    
    \item \textbf{Plug-and-play integration:} Our approach is \textbf{fully modular} and requires \textbf{no retraining or fine-tuning} of downstream classifiers, enabling seamless integration into existing vision systems and facilitating practical deployment.
\end{itemize}

\section{Background and Related Work}
\label{background}
Adversarial attacks introduce subtle, often imperceptible, perturbations to input data that can drastically alter a model's predictions, while remaining virtually undetectable to human observers. Remarkably, the human visual system demonstrates strong robustness to such perturbations, motivating researchers to explore biologically inspired mechanisms as potential defenses against adversarial threats.

\subsection{Foveal-Peripheral Vision and Multi-Resolution Processing}
Foveal-peripheral processing in primate vision, characterized by high acuity in the central fovea and progressively lower resolution in peripheral regions due to the non-uniform distribution of cones in the retina, has inspired significant research into improving neural network robustness. Additionally, multiscale filtering resulting from varying receptive field sizes across eccentricities in the primary visual cortex (V1) supports spatially non-uniform visual sampling\citep{voneblock}.

Previous works have emulated such biological mechanisms to enhance neural networks' resistance to small adversarial perturbations. For instance, \citep{rwrap} demonstrated that non-uniform sampling-based foveation improves robustness efficiently. However, its effectiveness diminishes with larger perturbations ($\epsilon > 1/255$). Similarly, VOneNet \citep{voneblock} integrated pretrained V1-simulating layers as a biologically constrained preprocessing step, significantly enhancing robustness to white-box attacks. However, this work is specifically designed for CNN based architecture. 

Approaches such as R-blur \citep{rblur}, which employs adaptive Gaussian blurring and color desaturation, and the Foveated Texture Transform (FTT) \citep{FTT}, further support the idea that inputs mimicking human retinal vision can  enhance the robustness of deep neural networks.. Notably, FTT improved robustness in scene classification tasks, where peripheral features often provide valuable contextual information. However, it showed limited effectiveness in object classification. We believe this is due to the crowd effect on the peripheral view which can be mitigated by saccade and multiple foveal sampling. While R-blur demonstrated relatively stronger robustness compared to aforementioned techniques, it still falls short compared to adversarially trained methods.

\subsection{Saccadic Eye Movements and Active Visual Exploration}
Saccadic eye movements—rapid, attention-guided shifts in gaze—enable primates to actively explore visual scenes by directing high-resolution foveal vision toward informative regions. This mechanism facilitates efficient scene understanding and context-aware perception.

Inspired by saccades, machine learning models employing hard attention or sequential observation have been developed. The Recurrent Models of Visual Attention (RAM) \citep{mnih2014recurrent} pioneered reinforcement learning-driven fixation point selection. Subsequent work extended these ideas, achieving notable successes in energy-efficient object recognition \citep{elsayed2019saccader,lukanov2021biologically}, object detection \citep{jaramillo2019foveated}, and low-power vision systems \citep{liu2023improved,rao2023dynamic}. These methods selectively sample and process visual information, significantly reducing computational requirements.

However, saccadic mechanisms remain relatively underexplored for adversarial robustness. Most existing biologically inspired defenses rely on static foveation or passive multi-resolution strategies. For example, although \cite{rblur} observed potential gains in robustness through manually selected fixation points, the development of an automated fixation point selection remains unexplored. This gap suggests a promising research direction: actively leveraging learned saccadic policies to strategically guide visual sampling could substantially enhance model resilience.

\subsection{Cortical Filling-in and Semantic Reconstruction}
The cortical filling-in phenomenon refers to the brain's ability to perceptually reconstruct missing or ambiguous visual input, maintaining coherent perception even when visual data is partially occluded or degraded. An well-known example is the physiological blind spot—despite the absence of photoreceptors in this region, humans perceive a continuous visual field by inferring missing information from surrounding context \citep{pessoa1998,komatsu2006}. Filling-in is crucial for recognizing occluded objects, preserving peripheral continuity, and forming comprehensive visual interpretations from limited or noisy inputs.

In computer vision, cortical filling-in principles underpin various generative tasks, such as image inpainting and masked image modeling. For instance, Context Encoders by \cite{pathak2016context} reconstruct missing image regions based on surrounding contextual information. More recently, Masked Autoencoders (MAE) \citep{he2022masked} have demonstrated effective reconstruction of masked visual inputs using transformer-based architectures.

Despite its fundamental role in human vision robustness, cortical filling-in remains largely underexplored in the context of adversarial defense. We hypothesize that filling-in mechanisms can significantly mitigate the effects of localized perturbations by internally reconstructing semantically coherent representations that are inherently more resistant to adversarial noise.
\section{Method}
\label{method}

\begin{figure*}
\centering
\includegraphics[scale=0.85]{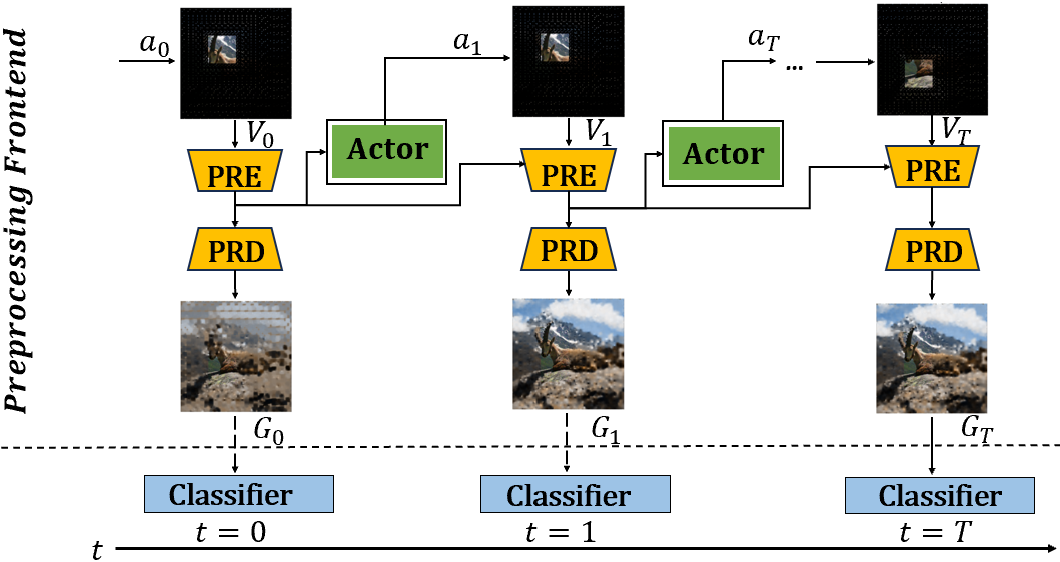}
\caption{Architecture overview. PRE (Predictive Reconstruction Encoder) comprises multiple ConvLSTM layers, and PRD (Predictive Reconstruction Decoder) is implemented as a convolutional neural network. The classifier can be any pretrained classfication model}
\label{architecture}
\end{figure*}

The overall architecture of the proposed SAFER-AiD consists of two core components: the predictive reconstruction module and the policy module. Figure \ref{architecture} illustrates the data flow between these components. The predictive reconstruction module receives a sequence of foveal-peripheral views of adversarially perturbed images. Its primary function is to infer missing information using prior knowledge and contextual cues, generating a perceptually consistent reconstruction of the scene to suppress adversarial perturbations. This process is supported by a Convolutional long short-term memory (LSTM), which maintains a hidden state, an embedding that captures the accumulated features from the sequence of glimpses observed thus far. 

The policy module is responsible for controlling saccadic movements. It uses the embedding generated by the predictive reconstruction module to strategically guide the foveal center toward perceptually informative regions. A new foveal-peripheral view is then sampled from the environment, and this iterative process continues until a termination condition is met.

\subsection{Foveal-peripheral sampling}

The foveal-peripheral sampling mechanism serves as an informative filter to emphasize semantically important regions while suppressing adversarial perturbations. The input image is first divided into an $N \times N$ grid of equally sized regions. During each sampling step, one region is selected as the foveal center and sampled at full resolution to preserve the semantic integrity of its content, while the remaining regions are treated as peripheral and sparsely sampled, enabling efficient suppression adversarial signals.

\begin{figure*}
\centering
\includegraphics[scale=0.40]{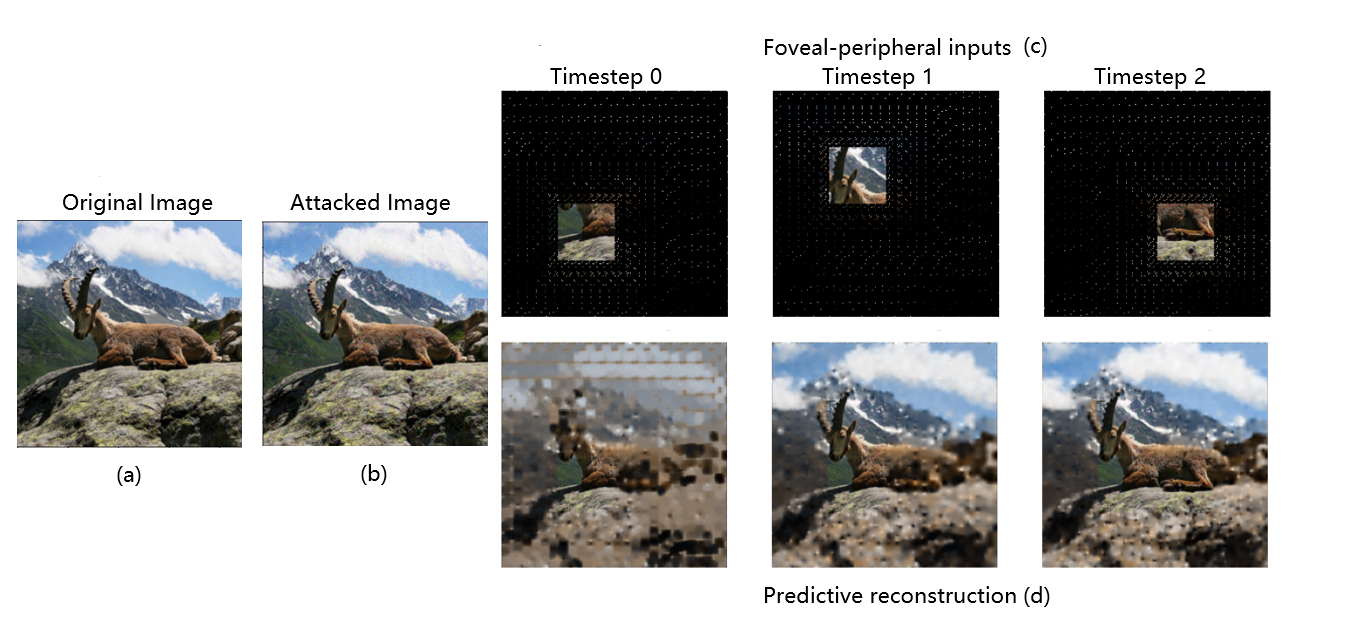}
\caption{An example of (a) original image, (b) attacked image, (c) three glimpses of foveal-peripheral sampled inputs, (d) reconstructed image after each glimpse.}
\label{goat}
\end{figure*}

Pixels within the foveal center are retained with probability 1, whereas those in the peripheral regions are sampled with a small probability $\mu$. The sampling probability for each pixel at location $(x, y)$ is defined as:
\begin{equation}
   Sample(x, y) =
   \left\{
   \begin{array}{ll}
   1 & \text(x, y) \text{ in foveal region} \\
   \mu & \text{otherwise}
   \end{array}
   \right.
\end{equation}
Using on these probabilities, a binary sampling mask $M \in \{0, 1\}^{X \times Y}$ is generated to determine which pixels are retained, with $X$ and $Y$ denoting the image dimensions. A new mask is independently generated for each glimpse, allowing the system to accumulate more contextual information over time. An example sequence of foveal-peripheral views is illustrated in Figure \ref{goat}(c).

\subsection{Predictive reconstruction model}

We adopt a multi-layer ConvLSTM architecture for the predictive reconstruction module (See Supplementary Material section A for model details) due to its strong capacity to learn spatiotemporal patterns. Detailed structure information of this module can be found in the Appendix. This module is trained using a self-supervised manner. Given the foveal-peripheral sampled inputs (as described in Section 3.1), it learns to progressively generate a perceptually consistent representation of the scene over time.
The training of the reconstruction model is independent of the downstream classifier and the saccade policy. During training, the foveal center is selected randomly to prevent sampling bias and to ensure that the model generalizes well across different saccade controllers.


To better align with human perceptual quality, we adopt the Local Structural Similarity Index (SSIM) \citep{1284395, 1292216} as our loss function. SSIM compares local image patches across three dimensions—luminance, contrast, and structural similarity—and is better aligned with human perception than pixel-wise losses. This makes it a more appropriate objective for our task, where perceptual quality and semantic fidelity are essential.

While our ultimate goal is to improve adversarial robustness, we do not train on adversarial examples. Instead, the foveal-peripheral sampling itself serves as a form of strong input corruption, making the reconstruction task inherently robust. Moreover, we observe that, during testing, the predictive nature of the reconstruction model tends to bypass the carefully crafted adversarial perturbations, producing outputs that closely resemble the original clean images. This behavior further enhancing robustness without requiring explicit adversarial training.


In this work, we train the predictive reconstruction model using a hybrid loss that combines mean squared error (MSE) and structural similarity (1–SSIM) as shown in Eq. \ref{EqLoss}. During the early stages of training, the model focuses primarily on minimizing MSE to stabilize learning and establish coarse structural alignment. As training progresses, we gradually increase the weighting factor $\lambda$ to balance the influence of SSIM, encouraging the model to learn finer structural details and local textures. In the later stages, SSIM becomes the dominant component of the loss, guiding the model toward producing perceptually sharper and more visually coherent renditions.
\begin{equation} \label{EqLoss}
\begin{split}
    Loss_{hybrid}(G_t,y) = \sum_{t=0}^{T} \big\{ & (1-\lambda) MSE(G_t, y) \\
    & + \lambda [1 - SSIM(G_t, y)] \big\}
\end{split}
\end{equation}

Figure \ref{goat} (d) gives an example of the sequence of reconstructed image from the corresponding foveal-peripheral views.



\subsection{Saccade Control with Advantage Actor-Critic model}

We formulate our problem as a multi-step episodic Markov Decision Process (MDP). The foveal-peripheral sampling scheme divides the original image  equally  into $N\times{N}$ non-overlapping patches which form the action space of saccade control $x_{ob} = \{x_0, x_1, ..., x_{N^2}\}$. Each selected foveal center plus its corresponding sparse peripheral view forms a single foveal-peripheral glimpse. 

The controller perceives the environment through the predictive reconstruction model. At each decision step $t$, we denote the LSTM's hidden state in the reconstruction model as $s_{1:t}$, which encodes a compressed representation of all foveal-peripheral glimpses received from time step 1 to $t$. This hidden state $s_{1:t}$ serves as the environment state used by the saccade controller  to choose the next action, $a_{t}\in x_{ob}$. We denote the saccade control policy as
 $\pi(s_{1:t};\theta _{p})\in[0,1]^{N^2}$, where $\theta_p$ represents the set of trainable parameters. The policy $\pi$ is a function that maps the environment  state, $s_{1:t}$, to a probability distribution over the patch sampling actions $a_{t}$.

The saccade controller is implemented and trained as an advantage actor critic (A2C) model. Both the actor and critic networks adopt the Resnet18 architecture. The controller is trained after the predictive reconstruction model has been fully trained; during this phase, the predictive reconstruction model remains frozen. The training follows the policy gradient method:
\begin{equation} \label{eq:3}
    \bigtriangledown_{\theta_p} J(\theta_p) =E[\sum_{t=1}^{T}\bigtriangledown_{\theta_p}log{\pi}(a_{t}|s_{1:t};\theta_p)A(s_{1:t},a_t)]
\end{equation}
where $A(s_{1:t},a_t)$ is the advantage function, calculated as the following:
\begin{equation}
    A(s_{1:t},a_{t}) = r_t + \gamma V(t+1) - V(t)
\end{equation}

The reward $r_t$ is application specific. Without loss of generality, we assume that the downstream application is image classification and accuracy is the evaluation metric. In this setting, $r_t$ is defined as the top-1 softmax score if the predicted label is correct, and as the negative top-1 softmax score otherwise, serving as a penalty. This reward design encourages the controller to select the foveal centers that improve the classification accuracy of the reconstructed image. 

$V(t)$ is the predicted value from critic model given the state vector $s_{1:t}$, and $V(t+1)$ represents the predicted value after agent take the action $a_t$. We train the saccade controller using the gradient of the advantage function rather than the absolute reward because the expected reward of sequence of random sampled actions may have a very large variance, leading to unstable training, while the advantage function $A(s_{1:t},a_t)$  helps to reduce the variance. 


To encourage the collection of novel information within a limited number of glimpses, the controller is restricted from selecting the same foveal center more than once. This is achieved by using an invalid action mask, $invalidMask\in{R^{N^2}}$. The $i$th entry of this vector is set to a value close to 0 if action $x_i$ corresponds to  a previously selected (i.e., invalid) location, and to 1 otherwise. The action probabilities are then adjusted element-wise using this mask as follows:

\begin{equation} \label{eq:8}
\begin{split}
    p'(a_{t}|s_{1:t};\theta_{p}) = \text{Softmax}\big[ & p(a_{t}|s_{1:t};\theta_{p}) \\
    & + \log(\text{invalid Mask}) \big]
\end{split}
\end{equation}
Equation \ref{eq:8} is differentiable and therefore can be integrated into the controller without disrupting the policy gradient flow described in Equation \ref{eq:3}. For each training and testing image, the initial  initial foveal center location  is  selected randomly.

\section{Evaluations}

In this section, we evaluate the adversarial robustness of SAFER-AiD in comparison to baselines. All experiments are conducted on adversarial samples generated from the ImageNet dataset. The images have a resolution of 224$\times$224 pixels and are categorized into 1000 classes. While we test SAFER-AiD with various downstream classifiers including multiple ViT variants and CNN-based architectures, we consistently use a foveal size of 56$\times$56 pixels. This is obtained by dividing each 224$\times$224 image into a 4$\times$4 grid, resulting in 16 non-overlapping patches. Additionally, we apply a peripheral sampling ratio of 6\% (See Supplementary material section B for a comparison between different sampling ratios) and 3 glimpses (See Supplementary material section C for a comparison between different number of glimpses).

\subsection{Improved Robustness to white-box attacks}
\begin{table*}
\centering
\small
\setlength{\tabcolsep}{10pt}
\renewcommand{\arraystretch}{1.0}
\begin{tabular}{p{5.5cm}ccc}
\toprule
\textbf{Model} & \multicolumn{3}{c}{\textbf{Top-1 Acc under Auto-PGD Attack}} \\
\cmidrule(lr){2-4}
& \textbf{1/255 [\%]} & \textbf{2/255 [\%]} & \textbf{4/255 [\%]} \\
\midrule
ResNet18                      & 0.0  & 0.0  & 0.0 \\
Rblur~Shah et al.~\cite{rblur}            & 18.0 & 2.7  & 0.0 \\
Rwrap~Vuyyuru et al.~\cite{rwrap}         & 0.0  & 0.0  & 0.0 \\
VOneBlock~Dapello et al.~\cite{voneblock} & 9.0  & 0.0  & 0.0 \\
SAFER-AiD\textsubscript{fix} (ResNet18)   & 23.3 & 14.6 & 6.2 \\
SAFER-AiD (ResNet18)                      & 28.6 & 17.5 & 8.8 \\
\midrule
ResNet50                      & 7.1  & 0.6  & 0.0 \\
AT~Wong et al.~\cite{AT2}                 & 49.6 & 42.4 & 30.8 \\
SAFER-AiD\textsubscript{fix} (ResNet50)   & 51.3 & 47.7 & 36.5 \\
SAFER-AiD (ResNet50)                      & 55.4 & 51.9 & 40.4 \\
\bottomrule
\end{tabular}
\caption{Model accuracy under varying Auto-PGD attack strengths. SAFER-AiD\textsubscript{fix} uses fixed foveal selection, while SAFER-AiD incorporates learned saccades.}
\label{whitebox}
\end{table*}
The non-differentiable nature of SAFER-AiD inherently defends against gradient-based adversarial attacks, as analytic gradients are not available. In Table \ref{whitebox}, to enable a fair comparison with existing bio-inspired defenses—such as Rblur\citep{rblur}, Rwrap\citep{rwrap}, and VOneBlock \citep{voneblock} —as well as the non-biological baseline of adversarial training, we manually select five fixation points for foveal vision. This approach  bypasses the non-differentiable saccade module, making it possible to conduct white-box attacks on our framework using Auto-PGD (APGD). We refer to this setting as "SAFER-AiD\textsubscript{$fix$}". 

Next, we evaluate the full system, including the reinforcement learning-based saccade policy, under the same APGD attack and refer to it as "SAFER-AiD". It is important to emphasize that the adversarial perturbs in this experiment are still generated without backpropagating gradients through the saccade module, due to its non-differentiable nature. As a result, the white-box attack scenario does not provide  a fair assessment of the saccade policy’s contribution to robustness. However, its effectiveness will be more appropriately evaluated in subsequent black-box attack scenarios.

For the first set of experiments, reported in the upper section of the table, we use ResNet-18 as the downstream classifier and evaluate robustness against Auto-PGD attacks with 25 steps. The evaluation is conducted under three adversarial budgets: $\epsilon = 1/255$, $2/255$, and $4/255$. These relatively small $\epsilon$ values are chosen because the three bio-inspired baselines, (Rblur, Rwrap, and VOneBlock), fail completely  at higher perturbation levels, achieving zero classification accuracy.

We also compare our framework to Adversarial Training\cite{AT2}, using ResNet-50 for both methods to ensure a fair comparison. Remarkably, our approach achieves higher robustness than the adversarially trained model. It is worth noting that, although the adversarial training baseline adopts a fast training strategy, it still requires approximately 12 hours on a 4×P100 machine. Moreover, this costly process must be repeated from scratch whenever a different classifier architecture is used. In contrast, our method involves a one-time training cost of approximately 10 hours for the predictive reconstruction model and 3 hours for the saccade policy, using a single NVIDIA RTX 2080 GPU. Once trained, our system requires no further training or fine-tuning, and it can be seamlessly integrated with any off-the-shelf classifier, offering both flexibility and efficiency.

\subsection{Improved Robustness to black-box attacks}

In this set of experiments, we evaluate the robustness of our system under more practical settings—transferable attacks and gradient free attack. We use four Vision Transformer (ViT) models as surrogate models to generate adversarial examples: ViT-B/16, PiT-B, CaiT-S/24, and Visformer-S. The target models include four additional ViT variants: DeiT-B, TNT-S, LeViT-256, and ConViT-B.

We evaluated the robustness of SAFER-AiD against two representative transferable attacks: Token Gradient Regularization (TGR) and MI-FGSM. For testing, we sample two images per class from the ImageNet validation set, resulting in a total of 2000 images. We ensure that all target models correctly classify the clean inputs; otherwise, it would be meaningless to measure the defense performance if models already misclassify clean examples. 

\begin{table*}
\centering
\begin{adjustbox}{width=\textwidth}
\renewcommand{\arraystretch}{0.98}
\begin{tabular}{llcccccccc|c}
\toprule
\multirow{2}{*}{\textbf{Surrogate models}} & \multirow{2}{*}{\textbf{}} 
& \multicolumn{8}{c|}{\textbf{Target Models}} & \multirow{2}{*}{\textbf{Mean}} \\
& & \textbf{ViT-B/16} & \textbf{PIT-B} & \textbf{CaiT-S/24} & \textbf{Visformer-S} 
  & \textbf{DeiT-B} & \textbf{TNT-S} & \textbf{LeViT} & \textbf{ConViT-B} & \\
\midrule
\multicolumn{11}{c}{\textbf{TGR}} \\
\midrule
\multirow{3}{*}{ViT-B/16} 
  & No Defense        & ---  & 47.9 & 15.0 & 44.3 & 14.2 & 24.9 & 41.5 & 13.4 & 28.7 \\
  & SAFER-AiD\textsuperscript{$-$}           & ---  & 53.5 & 42.2 & 50.2 & 43.4 & 48.6 & 50.8 & 37.1 & 46.5 \\
  & SAFER-AiD          & ---  & 57.4 & 46.0 & 54.0 & 47.1 & 52.4 & 55.2 & 40.3 & 50.3 \\
\midrule
\multirow{3}{*}{PIT-B}     
  & No Defense        & 27.6 & ---  & 10.7 & 0.5  & 11.4 & 6.8  & 6.5  & 10.2 & 10.5 \\
  & SAFER-AiD\textsuperscript{$-$}           & 35.9 & ---  & 19.0 & 16.0 & 25.8 & 12.8 & 14.8 & 17.3 & 20.2 \\
  & SAFER-AiD          & 40.6 & ---  & 23.1 & 20.0 & 30.0 & 17.0 & 19.0 & 21.6 & 24.6 \\
\midrule
\multirow{3}{*}{CaiT-S/24} 
  & No Defense        & 7.6  & 16.4 & ---  & 11.5 & 0.0  & 3.2  & 10.7 & 1.0  & 7.2 \\
  & SAFER-AiD\textsuperscript{$-$}           & 20.9 & 24.4 & ---  & 21.0 & 9.2  & 12.0 & 16.2 & 9.7  & 16.2 \\
  & SAFER-AiD          & 25.5 & 28.1 & ---  & 26.4 & 13.0 & 15.8 & 21.1 & 13.7 & 20.5 \\
\midrule
\multirow{3}{*}{Visformer-S} 
  & No Defense        & 42.4 & 14.8 & 21.4 & ---  & 23.7 & 12.7 & 16.1 & 24.5 & 22.2 \\
  & SAFER-AiD\textsuperscript{$-$}           & 48.3 & 21.4 & 33.6 & ---  & 37.7 & 25.0 & 22.7 & 34.5 & 31.9 \\
  & SAFER-AiD          & 53.4 & 25.6 & 37.5 & ---  & 41.9 & 29.3 & 26.4 & 39.1 & 36.2 \\
\midrule
\multicolumn{11}{c}{\textbf{MI-FGSM}} \\
\midrule
\multirow{3}{*}{ViT-B/16}
  & No Defense    & --- & 36.1 & 17.5 & 38.2 & 19.2 & 26.7 & 37.6 & 17.4 & 27.5 \\
  & SAFER-AiD\textsuperscript{$-$}          & --- & 40.1 & 30.6 & 42.5 & 30.3 & 35.2 & 41.4 & 27.4 & 35.4 \\
  & SAFER-AiD         & --- & 43.1 & 35.4 & 47.3 & 32.7 & 38.6 & 44.4 & 31.1 & 38.9 \\
\midrule
\multirow{3}{*}{PIT-B}
  & No Defense    & 44.1 & --- & 43.7 & 32.9 & 42.6 & 32.4 & 33.1 & 38.0 & 38.1 \\
  & SAFER-AiD\textsuperscript{$-$}          & 50.6 & --- & 48.2 & 38.3 & 48.1 & 41.3 & 39.4 & 45.6 & 44.5 \\
  & SAFER-AiD         & 53.3 & --- & 50.7 & 42.3 & 50.6 & 45.0 & 43.5 & 48.4 & 47.7 \\
\midrule
\multirow{3}{*}{CaiT-S/24}
  & No Dfense    & 14.3 & 28.0 & --- & 28.9 & 7.4 & 14.6 & 30.5 & 6.8 & 18.6 \\
  & SAFER-AiD\textsuperscript{$-$}          & 24.7 & 34.6 & --- & 32.5 & 20.2 & 24.7 & 32.9 & 20.8 & 27.2 \\
  & SAFER-AiD         & 29.5 & 39.6 & --- & 36.6 & 24.3 & 27.1 & 37.9 & 25.8 & 31.5 \\
\midrule
\multirow{3}{*}{Visformer-S}
  & No Defense    & 43.1 & 23.0 & 31.8 & --- & 33.3 & 25.6 & 25.4 & 28.9 & 30.2 \\
  & SAFER-AiD\textsuperscript{$-$}          & 51.0 & 30.3 & 40.5 & --- & 39.7 & 34.8 & 33.1 & 36.6 & 38.0 \\
  & SAFER-AiD         & 55.5 & 32.7 & 45.1 & --- & 43.9 & 38.5 & 35.9 & 41.0 & 41.8 \\
\bottomrule
\end{tabular}
\end{adjustbox}
\caption{Top-1 accuracy under transferable attacks (TGR and MI-FGSM). “SAFER-AiD\textsuperscript{$-$}” denotes the framework with random saccades, while “SAFER-AiD” uses learned saccades.}
\label{TGRMIFGSM}
\end{table*}

In Table \ref{TGRMIFGSM}, we report results under the TGR attack with parameters $\epsilon = 16/255$, $\alpha = 2/255$, and 10 attack iterations. Our method achieves an overall robustness improvement ranging from 9.0\% to 17.8\% with random saccades. When using the trained saccade policy, robustness further increases to 13.3\% to 21.6\%. In the rest part of the table, we evaluate under the MI-FGSM attack using $\epsilon = 16/255$, $\alpha = 2/255$, and 10 iterations. The proposed defense improves robustness by 6.4\% to 8.6\% with random saccades, and further boosts it to 9.6\% to 12.9\% with trained saccades.

\begin{figure*}
    \centering
    \begin{subfigure}[t]{0.32\textwidth}
        \centering
        \includegraphics[width=\linewidth]{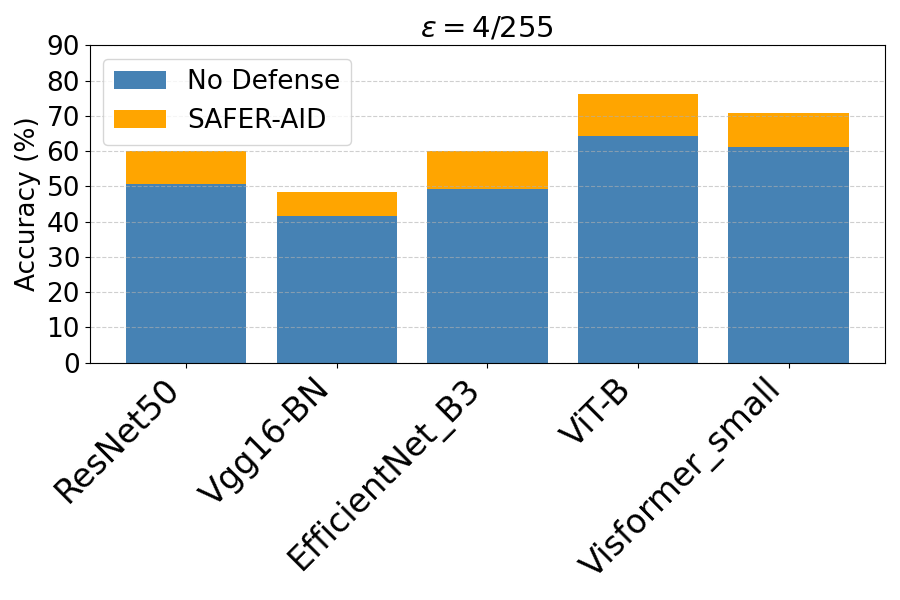}
        \caption{$\epsilon = 4/255$}
    \end{subfigure}
    \hfill
    \begin{subfigure}[t]{0.32\textwidth}
        \centering
        \includegraphics[width=\linewidth]{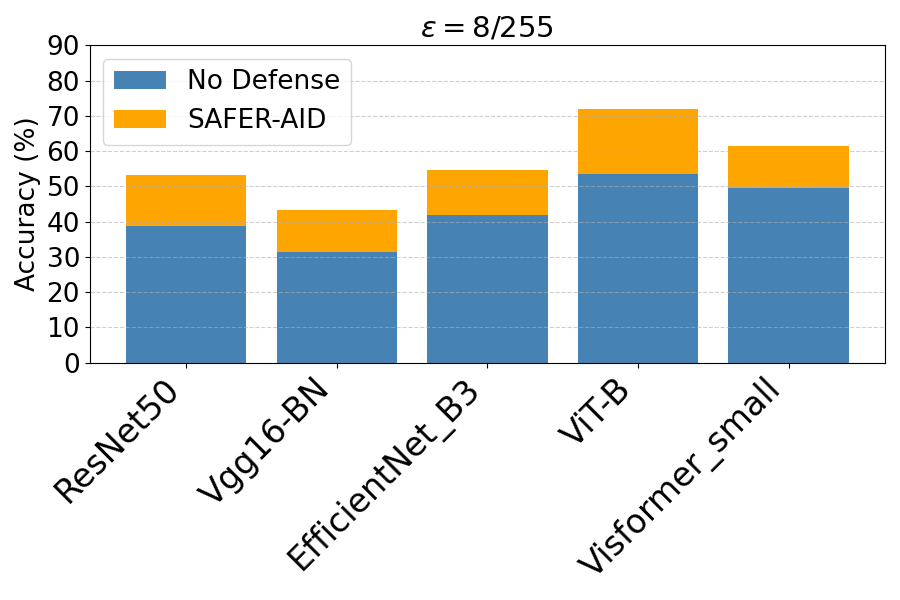}
        \caption{$\epsilon = 8/255$}
    \end{subfigure}
    \hfill
    \begin{subfigure}[t]{0.32\textwidth}
        \centering
        \includegraphics[width=\linewidth]{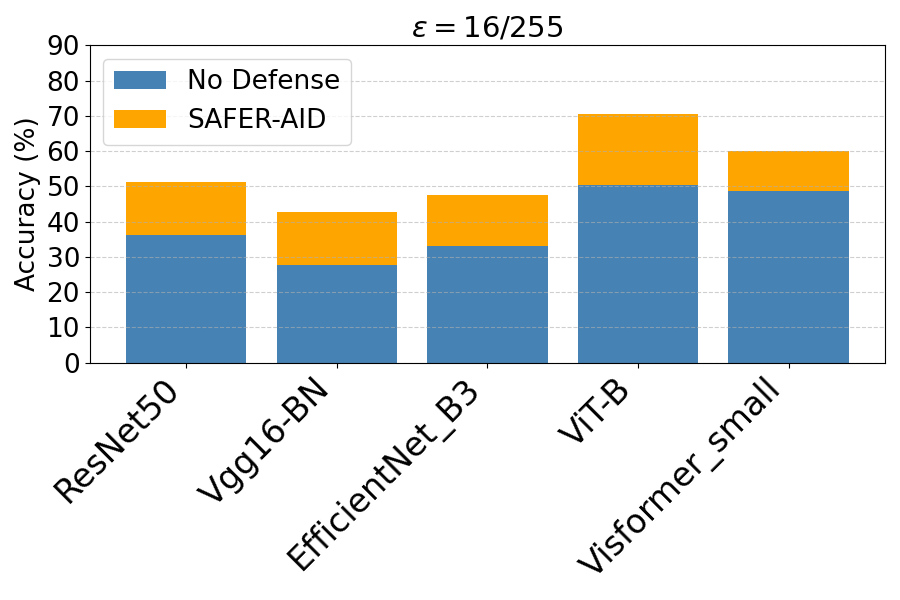}
        \caption{$\epsilon = 16/255$}
    \end{subfigure}
    \caption{Top-1 accuracy under adversarial perturbation with different $\varepsilon$.}
    \label{fig:spsa}
\end{figure*}

Another black-box attack we evaluated is Simultaneous Perturbation Stochastic Approximation (SPSA) \citep{uesato2018adversarial}, a gradient-free optimization method that approximates the gradient by repeatedly querying the model with randomly perturbed inputs. Figure \ref{fig:spsa} shows the improvement that SAFER-AiD can bring to some typical CNN and ViT models under the SPSA attack at perturbation strengths of $\epsilon = {4/255,\ 8/255,\ 16/255}$ with a step size $\alpha = 2/255$. The results demonstrate that our biologically inspired vision system also maintains robustness against gradient-free attacks like SPSA. Specifically, when $\epsilon = 4/255$, the average Top-1 accuracy of the six undefended models is 46.7\%, while applying our framework raises it to 56.5\%. At $\epsilon = 8/255$, the accuracy increases from 36.4\% (undefended) to 50.4\% (defended), and at $\epsilon = 16/255$, from 32.7\% to 47.9\%.

Compared to undefended system, our system exhibits smaller performance drop as the attack strength increases. While the undefended models experience an average accuracy drop a 14.0\%  when the perturbation budget increases from $\epsilon = 4/255$ to $\epsilon = 16/255$, our method exhibits only  an 8.6\% reduction. This demonstrates that our framework not only improves baseline robustness but also mitigates performance degradation under stronger perturbations.
\subsection{Predictive reconstruction is a faithful perceptual rendition}

\begin{figure}[h]
  \centering
  \includegraphics[width=1\linewidth]{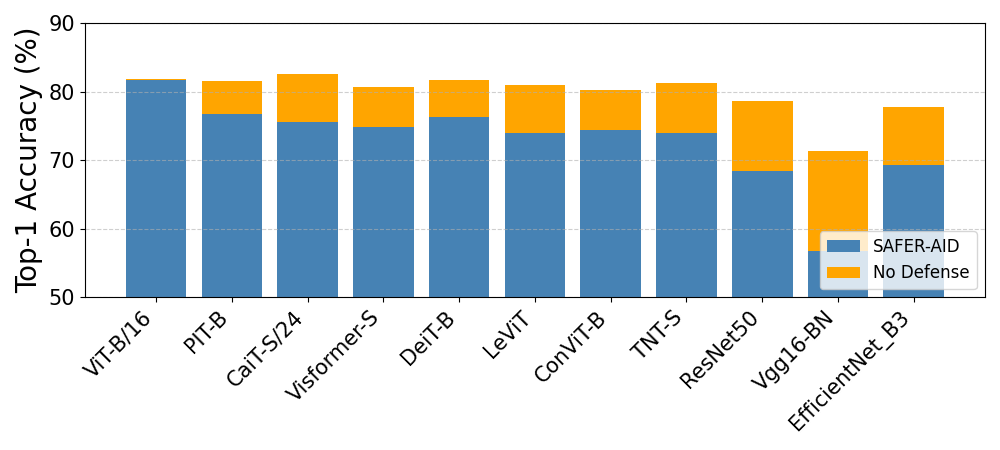} 
  \caption{Accuracy comparison with clean inputs.}
  \label{fig:bar_chart}
\end{figure}

The human visual system maintains robust perception without sacrificing  accuracy under clean (non-adversarial) conditions. In this experiment, we assess  whether integrating our framework results in any significant degradation  in classification performance on clean, unperturbed images.

We evaluate SAFER-AiD with downstream classifiers based on eight different Vision Transformer (ViT) variants and three CNN-based architectures, covering a broad range of commonly used model families. As shown in Figure~\ref{fig:bar_chart}, SAFER-AiD demonstrate the highest compatibility with ViT-based downstream classifiers. When using the ViT-Base model as the classifier, our system shows virtually no degradation in performance on clean images. This may be attributed to the attention mechanism in ViTs, which can help suppress noise introduced by foveal-peripheral sampling and the reconstruction process. However, as demonstrated in earlier experiments, ViTs alone remain highly vulnerable to adversarial attacks. A truly robust vision system benefits from the combined effects of saccadic foveal-peripheral viewing, cortical filling-in, and attention-driven perception. 

While CNN-based classifiers experience a slightly greater accuracy degradation for clean images when used with SAFER-AiD, they gain substantial improvements in adversarial robustness as reported in previous sections. 
\section{Conclusion and future directions}
In this work, we propose a novel defense framework that integrates three core mechanisms from human vision—foveal-peripheral sampling, saccadic eye movements, and cortical filling-in—to enhance adversarial robustness. Inspired by the brain's ability to perform sparse, goal-directed sensing and predictive reconstruction, our approach consistently improves robustness across CNNs and ViTs, outperforming prior biologically inspired defenses and adversarial training baselines.

A key advantage is its plug-and-play design, requiring no retraining or fine-tuning of downstream classifiers, allowing seamless integration into existing vision systems. Our results validate the hypothesis that non-uniform sparse sampling and predictive coding are critical for robust perception, enabling semantic reconstruction from sparse and adversarial inputs.

Looking forward, we aim to extend this framework to dynamic settings like video understanding and object tracking, and explore hardware-efficient implementations for real-time, low-power deployment.
\newpage
{
    \small
    \bibliographystyle{ieeenat_fullname}
    \bibliography{main}
}

\end{document}